\documentclass[letterpaper, 10 pt, journal, twoside]{ieeetran}

\IEEEoverridecommandlockouts                              

\usepackage[utf8]{inputenc} 
\usepackage[T1]{fontenc}    
\usepackage[colorlinks]{hyperref}       
\usepackage{url}            
\usepackage{booktabs}       
\usepackage{amsfonts}       
\usepackage{nicefrac}       
\usepackage{microtype}      

\usepackage[utf8]{inputenc} 
\usepackage[T1]{fontenc}    
\usepackage{hyperref}       
\usepackage{booktabs}       
\usepackage{amsfonts}       
\usepackage{nicefrac}       
\usepackage{microtype}      

\usepackage{amsmath}
\usepackage{amssymb}
\usepackage{multirow}
\usepackage{multicol}
\usepackage{color}
\usepackage{wrapfig}
\usepackage{xcolor}
\usepackage{xspace}
\usepackage{caption}
\usepackage{tablefootnote}
\usepackage{epsfig}
\usepackage{makecell}
\usepackage{tikz-cd}
\usepackage{graphicx}

\usepackage{adjustbox}
\usepackage{array}
\usepackage{wrapfig}
\usepackage{bbm}
\usepackage{tabularx}
\usepackage{subcaption}
\usepackage{gensymb} 
\usepackage[backend=bibtex,minbibnames=5,maxbibnames=5,firstinits=true]{biblatex}

\usepackage{multicol}
\usepackage{animate}
\usepackage{tikz}
\usetikzlibrary{positioning, fit, arrows.meta, shapes}

\newcommand{\inv}{^{\raisebox{.2ex}{$\scriptscriptstyle-1$}}}
\newcommand\onedot{.\enspace}
\def\eg{\emph{e.g.,}\enspace} 
\def\ie{\emph{i.e.,}\enspace}

\def\etc{\emph{etc}\onedot}

\def\etal{\emph{et al}\onedot}

\newcommand{\myparagraph}[1]{\vspace{0.0em}\noindent\textbf{#1}}
\newcommand\red[1]{\textcolor{red}{#1}} %
\newcommand\new[1]{#1} %

\title{\LARGE \bf FoV-Net: Field-of-View Extrapolation Using Self-Attention and Uncertainty}

\author{Liqian Ma$^{1}$ \quad Stamatios Georgoulis$^{2}$ \quad Xu Jia$^{3}$  \quad Luc Van Gool$^{1,2}$
}

\begin{document}


\twocolumn[{
\renewcommand\twocolumn[1][]{#1}
\maketitle
\begin{center}
\centering
  \includegraphics[width=1\linewidth]{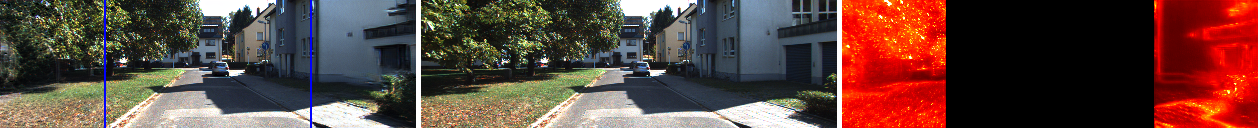}
\captionof{figure}{Given the present and past frames with narrow FoV, we hallucinate the present frame with wide FoV (left: hallucination, middle: ground truth) and predict the associated uncertainty (right). The past frames are \new{omitted} for brevity.}
\label{fig:teaser}
\end{center}
}]

\begin{abstract}

The ability to make educated predictions about their surroundings, and associate them with certain confidence, is important for intelligent systems, like autonomous vehicles and robots. It allows them to plan early and decide accordingly. Motivated by this observation, in this paper we utilize information from a video sequence with a narrow field-of-view to infer the scene at a wider field-of-view. To this end, we propose a temporally consistent field-of-view extrapolation framework, namely FoV-Net, that: (1) leverages 3D information to propagate the observed scene parts from past frames; (2) aggregates the propagated multi-frame information using an attention-based feature aggregation module and a gated self-attention module, simultaneously hallucinating any unobserved scene parts; and (3) assigns an interpretable uncertainty value at each pixel. Extensive experiments show that FoV-Net does not only extrapolate the temporally consistent wide field-of-view scene better than existing alternatives, but also provides the associated uncertainty which may benefit critical decision-making downstream applications. 
Project page is at \url{http://charliememory.github.io/RAL21_FoV}.

\end{abstract}


{
  \renewcommand{\thefootnote}%
    {\fnsymbol{footnote}}
  \footnotetext{Manuscript received: October 14, 2020; Revised: January 17, 2021; Accepted: February 19, 2021. 
  This paper was recommended for publication by Editor Dana Kulic upon evaluation of the Associate Editor and Reviewers' comments.
  This work was supported by Toyota Motors Europe.
  
  $^{1}$ L. Ma and L. Van Gool are with with the Department of Electrical Engineering, KU Leuven, Belgium {\tt\footnotesize \{liqian.ma,luc.vangool\}@esat.kuleuven.be}
  
  $^{2} $ S. Georgoulis and L. Van Gool are with the Department of Information Technology and Electrical Engineering, ETH Zürich, Switzerland {\tt\footnotesize \{georgous,vangool\}@vision.ee.ethz.ch}
  
  $^{3} $ X. Jia is with the School of Artificial Intelligence, Dalian University of Technology, China {\tt\footnotesize xjia@dlut.edu.cn}
  
  Digital Object Identifier (DOI): see top of this page.
}
}

\section{Introduction}
\label{sec:intro}
\IEEEPARstart{I}{n} our pursuit of intelligent machine perception, it is crucial to endow systems, like autonomous vehicles and robots, with an awareness of the scene content beyond their immediately visible field-of-view (FoV). Simply put, the system should be able to hallucinate its surroundings, and associate each prediction with certain confidence, which could help it plan early and decide accordingly.
\new{For example, when a moving camera is turning right at a corner in a road-bound scene, the right blind part is completely unobserved but can be reasonably hallucinated. Or when observing a car in a neighboring lane over time, wide FoV synthesis can help to reason about its future position, also beyond its immediately visible FoV. Such wide FoV hallucination ability can benefit vision-based navigation~\cite{wang2019reinforced}, exploration~\cite{jayaraman2018learning}, and augmented-reality telerobotics system~\cite{makhataeva2020augmented}.}
To realize this idea, we can draw inspiration from how humans use vision to relate themselves to the world around them. Humans clearly have a situational awareness that goes beyond their narrow FoV. On the one hand, this is grounded in a capability to propagate local scene content from past observations (\new{\eg} anticipate the future position of a previously observed building based on the car's trajectory when driving). On the other hand, it is due to an ability to hallucinate global scene content for unobserved regions based on the scene's context (\new{\eg} the unobserved side views in a driving scene are likely to contain trees if the car crosses a forest area). Most importantly, humans can typically assign a degree of confidence in these judgments to quantify their intuition.

\begin{figure}[t]
\centering
  \includegraphics[width=1\linewidth]{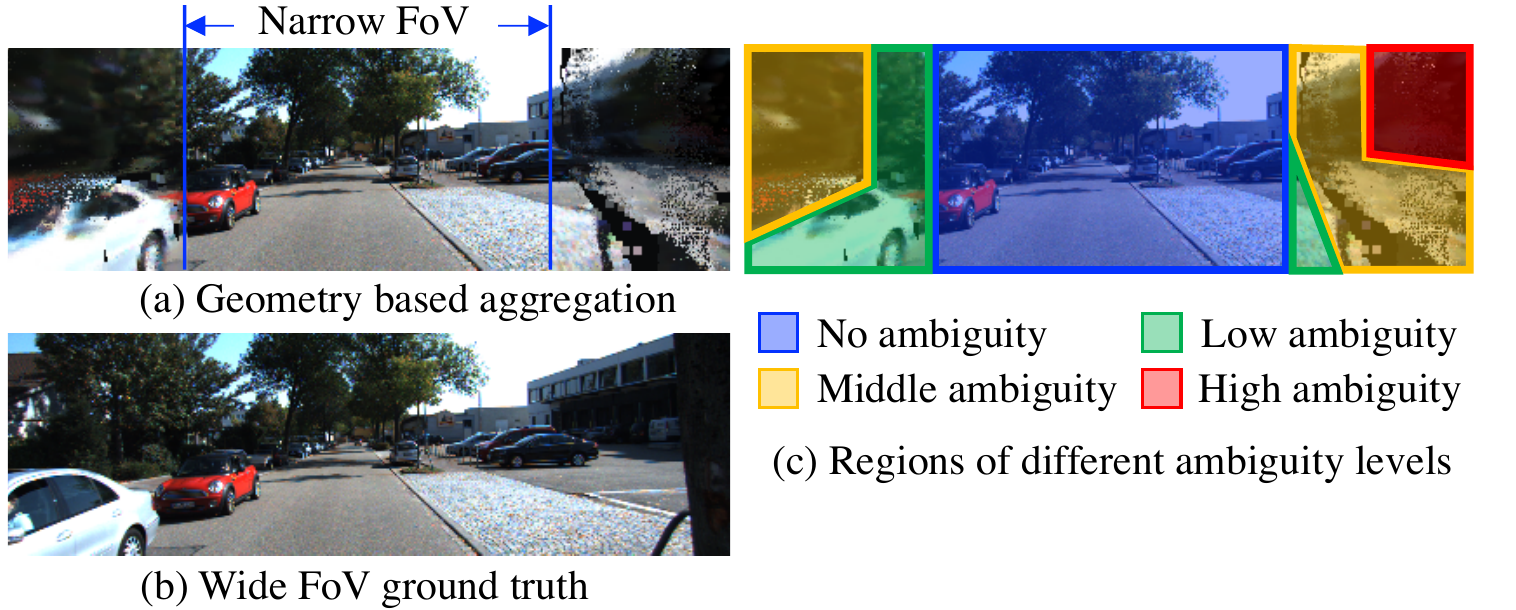}\\\
    \caption{\new{Ambiguity illustration. Best viewed in the digital form.} 
    }
  \label{fig:ambiguity}
\end{figure}

Motivated by these observations, in this paper, we tackle the problem of \textit{FoV extrapolation}. The goal is to leverage information from a video sequence with narrow FoV (including the present and few past frames) in order to infer the (present) scene at a wider FoV (see Fig.~\ref{fig:teaser}). There are several challenges associated with this problem. (1) A large image size discrepancy between the input narrow FoV frames and the output wide FoV frame has to be bridged, and the results should be temporally consistent. (2) Some areas in the wide FoV frame may change significantly or even not appear in any of the past narrow FoV frames. For example, far away objects in the past narrow FoV frames need upscaling or novel view synthesis, and some occluded or unobserved regions need to be inpainted. Thus, lots of details need to be hallucinated in the wide FoV frame. 
(3) There is ambiguity existing between the narrow FoV observations and the wide FoV ground truth.
The ambiguity mainly comes from two sources: the unobserved information and possible 3D estimation errors.
In particular, the pixels in the wide FoV frame can be roughly divided into four types (see Fig.~\ref{fig:ambiguity}): (a) the observed narrow FoV pixels in the present frame (no ambiguity); (b) the propagated pixels from past frames with accurate propagation (low ambiguity); (c) the propagated pixels from past frames with noisy propagation (medium ambiguity); (d) the unobserved regions (high ambiguity). When the ambiguity is high, strong enforcement of pixel reconstruction may mislead the training process. In contrast, perceptual and adversarial losses can be more suitable to predict a plausible outcome.

\begin{figure*}[t]
  \centering
  \includegraphics[width=1\linewidth]{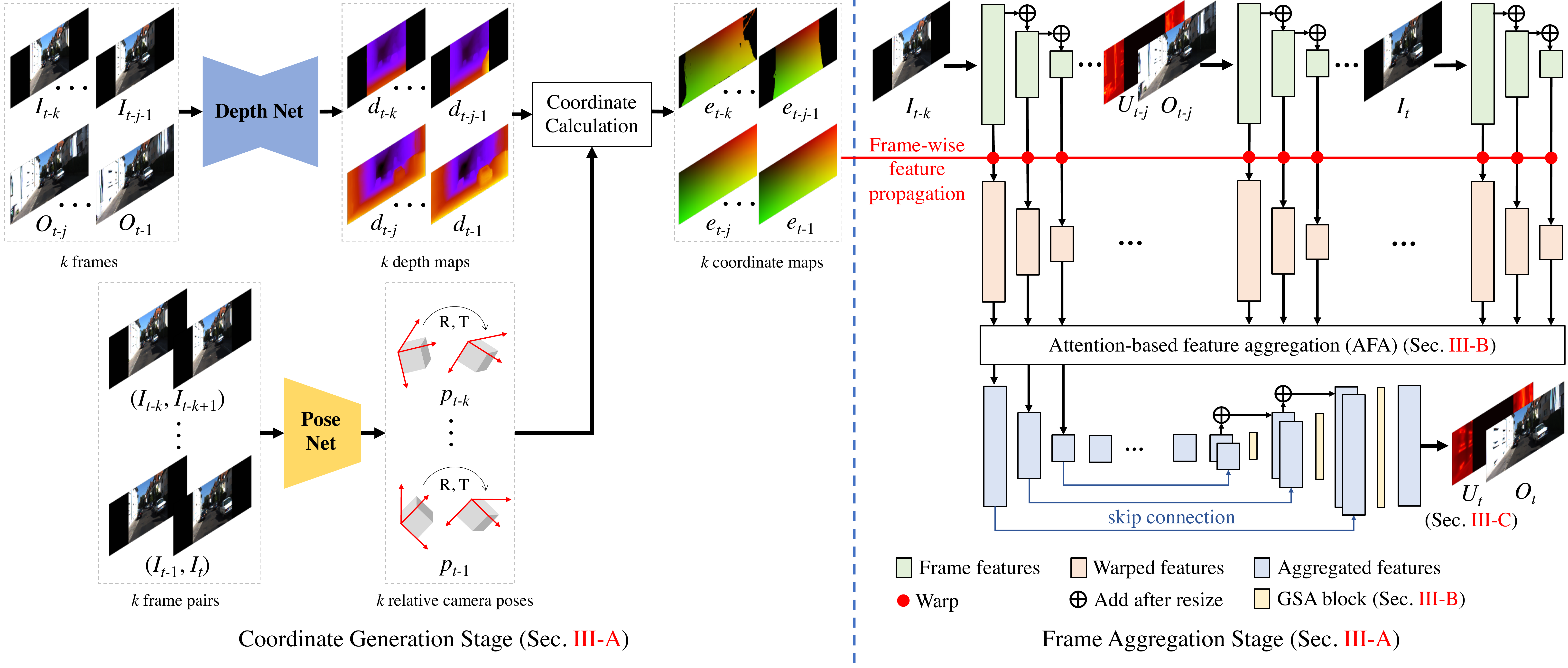}\\
     \caption{The proposed \textit{FoV-Net} framework. Left: the coordinate generation stage (Sec.~\ref{sec:overview}), which estimates the scene-level rigid flow, governed by the camera motion, and uses it to calculate the coordinates (\new{\ie} pixel displacements). Right: the frame aggregation stage (Sec.~\ref{sec:overview}), which utilizes the generated coordinates to propagate past frames information on a multi-scale feature level -- denoted as \red{red} dots -- and then aggregates the propagated features with an Attention-based Feature Aggregation module (Sec.~\ref{sec:sa}). To synthesize the final result $O_t$, a U-Net architecture is adopted to in/out-paint the missing regions, where a Gated Self-Attention module (Sec.~\ref{sec:sa}) is introduced to handle different ambiguities for better generation quality. Concurrently, an uncertainty map $U_t$ is jointly estimated to interpret the hallucination uncertainty at each pixel and guide the learning by reducing supervision ambiguity (Sec.~\ref{sec:uncertainty}). For $I_t$, we use the identical coordinates as $e_t$, namely, the features are not changed after warping. The coordinate $e_i$ is also used in the GSA blocks to warp the past hidden states, but we omit the arrows here. The discriminator networks \new{for adversarial losses} are also omitted for clarity. Due to FoV-Net's recurrent nature, note that previous outputs ${\{O_{t-i}\}}_{i=1,\dots,j}$ and ${\{U_{t-i}\}}_{i=1,\dots,j}$ become future inputs ${\{I_{t-i}\}}_{i=1,\dots,j}$ for temporal coherency purposes. \new{Best viewed in the digital form.}
     }
  \label{fig:framework}
\end{figure*}

To address these challenges, we propose a temporally consistent FoV extrapolation framework called \textit{FoV-Net}, which consists of two stages (see Fig.~\ref{fig:framework}). A coordinates generation stage propagates past narrow FoV frames into the present wide FoV frame by leveraging 3D scene information (addressing challenge 1). A frame aggregation stage combines the multi-frame propagated information, simultaneously hallucinating fine details and unobserved scene parts (addressing challenges 2\&3). Specifically, in the frame aggregation stage, we introduce an Attention-based Feature Aggregation (AFA) module to better fuse the propagated multi-frame information on the feature level, and a Gated Self-Attention (GSA) module to handle the discussed ambiguities and improve the generation quality (addressing challenges 2\&3). Finally, we introduce an uncertainty mechanism to interpret the hallucination uncertainty at each pixel and guide the learning by reducing supervision ambiguity (addressing challenge 3). Such hallucination uncertainty is rarely discussed in the image synthesis field, but is quite important for practical downstream decision-making applications. Fig.~\ref{fig:teaser} gives an example of our FoV extrapolation with its associated uncertainty. 

\section{Related Work}
\label{sec:related}

\myparagraph{Video-based image synthesis.}
This problem exists in various forms in the literature, including video inpainting, video extrapolation, novel view synthesis, future video prediction, video-to-video synthesis, \etc Video inpainting~\cite{huang2016temporally,wang2019video,kim2019deep,chang2019free,LGTSM,Gao-ECCV-FGVC} aims to hallucinate the missing pixels through warping or generate the missing pixels conditioned on the neighboring (in spatial or temporal dimensions) visible pixels. The typical setting is to utilize the past, present, and future frames to inpaint the missing pixels in the present frame, all within narrow FoV. Video extrapolation~\cite{lee2019video,choi2020deep,zhang2014parallax,guo2016joint} usually adopts 2D or 3D geometry-based image warping and stitching techniques to blend the observed pixels of adjacent narrow FoV frames in order to extend the FoV, but totally ignores any unobserved pixels and object view changes. Novel view synthesis~\cite{park2017transformation,sun2018multi,hedman2018deep,choi2019extreme,chen2019nvs,flynn2019deepview} aims to generate images of a given object or scene from different viewpoints by blending the observed pixels, as well as hallucinating a few missing pixels mainly for dis-occlusion. When applied to scenes, it is heavily reliant on highly accurate multi-view geometry to produce good results. Future video prediction~\cite{denton2018stochastic,byeon2018contextvp,liu2017voxelflow,lotter2016deep,gao2019disentangling,liu2018dyan} focuses on hallucinating future frames conditioned on the past and present frames, all within narrow FoV. Undoubtedly, this task entails higher uncertainty in the predictions. Video-to-video synthesis~\cite{wang2018video,chen2017coherent,shi2016real,bansal2018recycle} mainly transfers the appearance while preserving the structure of the input (\new{\eg} semantic maps). Thus the input and output are usually well aligned and within narrow FoV.
\new{Unlike} the existing forms, our goal is to infer the present scene at a wider FoV, including the observed and unobserved pixels, conditioned on the past and present narrow FoV observations. 
\new{Note that, when the camera is moving forward, most out-of-view regions of the present narrow FoV frame are actually future predictions as far as observations in the past frames are concerned. While if the camera is turning around at a corner, part out-of-view regions may be totally unobserved before, which makes the problem more challenging.}
The object size and view may change significantly, leading to large unobserved regions. 
To the best of our knowledge, none of the existing video-based image synthesis forms fully covers the needs of our intricate problem. 

\myparagraph{Attention.}
The success of self-attention models in natural language processing has inspired various applications in the computer vision field, such as in image recognition~\cite{hu2019local,SAN}, image synthesis~\cite{SAGAN,brock2018large,parmar2018image}, video prediction~\cite{jia2016dynamic,wang2018non}, and imitation learning~\cite{icra2020attentive}. Self-attention can be formulated as locally adaptable convolutional layers
with different weights for different types of image regions~\cite{SAN}. In our FoV extrapolation problem, we observe that such local adaptability is essential in terms of hallucination quality, since different image regions have different characteristics (see Fig.~\ref{fig:ambiguity}). To improve the hallucination quality, we propose a novel gated self-attention module, motivated by the success of gated convolution in image inpainting~\cite{deepfillv2} and that of self-attention in image recognition~\cite{SAN}. 
Note that, the gated self-attention concept is not new and has been introduced in natural language processing~\cite{dhingra2016gated,tran2017generative,zhao2018paragraph,lai2019gated}. However, in this work, we extend it to the video domain. 
Additionally, in order to better aggregate the propagated multi-frame information, we propose an Attention-based Feature Aggregation (AFA) module -- which is not to be confused with self-attention used above -- that makes our framework more robust to propagation errors and improves the generation quality (see Sec.~\ref{sec:sa}).

\myparagraph{Uncertainty estimation.}
Reasoning about \new{the} uncertainty of neural network prediction is essential for practical decision-making applications~\cite{loquercio2020general}. Although uncertainty estimation has been proved to be effective for several computer vision tasks, including object detection~\cite{he2019bounding}, semantic segmentation~\cite{multitaskUncertainty,huang2018efficient}, depth estimation~\cite{depthUncertainty,yang2020d3vo} and optical flow~\cite{flowUncertainty}, it remains largely unexplored in the image synthesis literature. In this paper, we propose a hallucination uncertainty estimation mechanism, which not only enables the prediction of uncertainty, but also guides the learning by reducing supervision ambiguity. In general, there are different ways to estimate the uncertainty, including empirical estimation, predictive estimation, and Bayesian estimation (see ~\cite{depthUncertainty,loquercio2020general} for a comprehensive survey). Among them, the predictive estimation is \new{desirable} due to its effectiveness and efficiency, and has been explored in several computer vision tasks~\cite{depthUncertainty,flowUncertainty,multitaskUncertainty}. As a positive side effect, when integrated with the training objective in our problem, the predictive uncertainty can naturally weight the loss functions spatially to reduce the supervision ambiguity.

\section{FoV-Net framework}
\label{sec:framework}

\subsection{System overview}
\label{sec:overview}
Given a present narrow FoV frame $I_t$ and $k$ past narrow FoV frames ${\{I_{t-i}\}}_{i=1,\dots,k}$ \new{($k=5$ in our experiment)}, our goal is to synthesize the present wide FoV frame $O_t$ -- close to ground truth $W_t$ -- and predict the hallucination uncertainty $U_t$. In addition, the adjacent synthesized results $O_{t-1}$ and $O_{t}$ should be temporally consistent. To achieve these, we propose a two-stage recurrent framework (Fig.~\ref{fig:framework}) consisting of a coordinate generation stage and a frame aggregation stage, coupled with a hallucination uncertainty mechanism (Sec.~\ref{sec:uncertainty}). The coordinate generation stage is designed to estimate the scene-level rigid flow, governed by the camera motion, and use it to generate coordinates (\new{\ie} pixel positions) in order to spatially propagate information from the past \new{narrow FoV} frames. The frame aggregation stage is designed to aggregate the past narrow FoV frames $I_{t-k},\dots,I_{t-1}$ and present narrow FoV frame $I_{t}$ into one wide FoV image, as well as hallucinate the unobserved missing regions. To enforce temporal coherency, we use a simple recurrent feed-forward strategy: replace the narrow FoV inputs ${\{I_{t-i}\}}_{i=1,\dots,j}$ with the previous outputs ${\{O_{t-i}\}}_{i=1,\dots,j}$, and feed the corresponding previous uncertainty ${\{U_{t-i}\}}_{i=1,\dots,j}$ by channel-wise concatenation ($j=2$ in our settings). We analyze each stage of our framework below.

\myparagraph{Coordinate generation stage}, that builds upon Monodepth2~\cite{monodepth2}, consists of a depth network $\mathcal{D}_{\theta_\mathcal{D}}$ and a relative camera pose network $\mathcal{P}_{\theta_\mathcal{P}}$\footnote{The subscripts $\{\theta_\mathcal{D}, \theta_\mathcal{P}, \theta_\mathcal{A}, \theta_\mathcal{Q}, \theta_\mathcal{T}\}$ are network parameters. We regularly omit the subscripts for brevity.}. During training, \new{$\mathcal{P}$} takes a pair of two \new{temporally} adjacent frames as input and outputs the relative camera pose, and \new{$\mathcal{D}$} takes one frame of the pair as input and outputs its depth. During inference, however, we do not have access to the wide FoV frame which is required by the inverse warping operation~\cite{jaderberg2015spatial} in order to propagate the pixels between two frames.
To address this, we design a forward warping strategy to propagate the past narrow FoV frames to the present wide FoV frame. We first utilize the depth maps from the past narrow FoV frames to calculate the rigid flow $f^{rig}_{t\rightarrow i}(\hat{e}_{t\rightarrow i})$ from the present frame $I_t$ to the past frame $I_i$, using Eq.~\ref{eq:rigid_flow}. 
\begin{align}
    f^{rig}_{t\rightarrow i}(\hat{e}_{t\rightarrow i})=K\mathrm{T}_{i\rightarrow t}\mathcal{D}_i(c_i)K\inv c_i - c_i,
    \label{eq:rigid_flow}
\end{align}
where $K$ denotes the camera intrinsic matrix, $\mathrm{T}_{i\rightarrow t}$ denotes the relative camera pose, and $c_i$ denotes homogeneous coordinates of pixels in frame $I_i$. 
Then, using the calculated flow, we compute the spatial mapping, \new{\ie} coordinate map $\hat{e}_{t\rightarrow i}$, that \new{spatially matches the pixel positions of present frame $I_t$ to the corresponding ones of past frame $I_i$}. Finally, we reverse this correspondence, \new{\ie} $e_i=\text{reverse}(\hat{e}_{t\rightarrow i})$, which now corresponds to the spatial positions from the past frame $I_i$ to the present frame $I_t$. This coordinate map $e_i$ will be used to propagate features in the frame aggregation stage using bilinear sampling. 

\myparagraph{Frame aggregation stage} is designed to aggregate the past narrow FoV frames $I_{t-k},\dots,I_{t-1}$ and present narrow FoV frame $I_{t}$ as well as previous wide FoV results ${\{O_{t-i}\}}_{i=1,\dots,j}$, ${\{U_{t-i}\}}_{i=1,\dots,j}$, into one wide FoV image, simultaneously hallucinating unobserved missing regions. It contains aggregation network $\mathcal{A}_{\theta_\mathcal{A}}$, image discriminator network $\mathcal{Q}_{\theta_\mathcal{Q}}$, and temporal discriminator network $\mathcal{T}_{\theta_\mathcal{T}}$. The aggregation network $\mathcal{A}$ first extracts a residual multi-scale feature pyramid (with $N=3$ levels) from each frame using an encoder with shared weights, and propagates the multi-scale features using the computed coordinates $e_i$. Then, an Attention-based Feature Aggregation (AFA) module (Sec.~\ref{sec:sa}) aggregates the propagated features. To synthesize the final result based on the aggregated features, a U-Net decoder is designed, where we introduce the Gated Self-Attention (GSA) module (Sec.~\ref{sec:sa}) to adaptively handle ambiguities and improve the generation quality.

\subsection{Self-attention mechanism}
\label{sec:sa}

\begin{wrapfigure}{l}{0.25\textwidth}
\centering
  \includegraphics[width=1\linewidth]{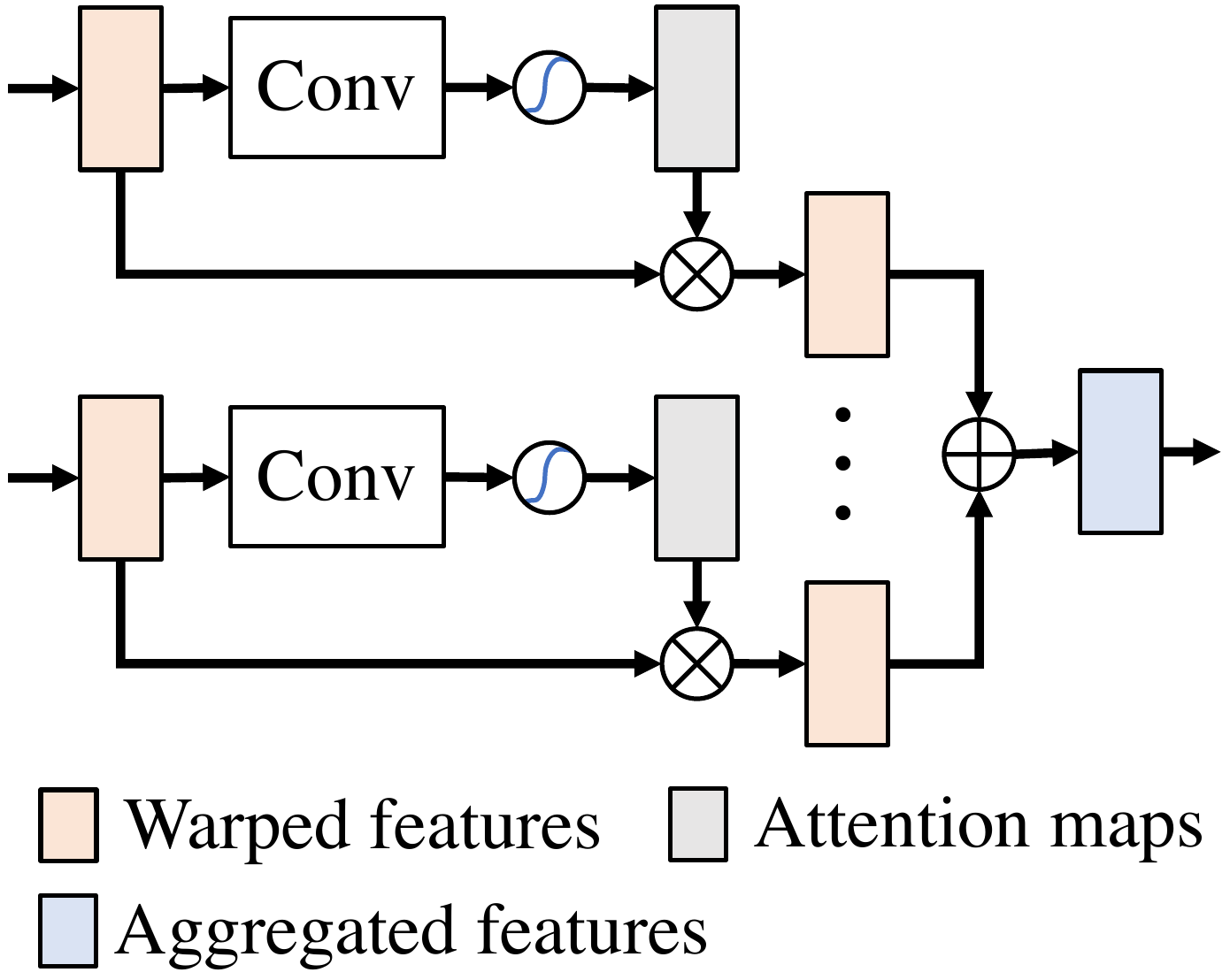}\\
    \caption{AFA module. Same color coding as Fig.~\ref{fig:framework}}
  \label{fig:attn_agg_module}
\end{wrapfigure}

\textbf{Attention-based Feature Aggregation (AFA).} In Fig.~\ref{fig:attn_agg_module}, to aggregate the propagated features, each set of propagated \new{multi-scale features maps (Fig.~\ref{fig:framework} right)} are fed into a convolution layer followed by softmax normalization to predict frame-wise spatial attention maps (\new{\ie} one  
channel attention map for each frame). 
Then, the propagated feature maps are multiplied by the attention maps and summed across all frames. This attention-based aggregation module can learn to select the useful features among these frames to address the issues caused by depth/pose errors and frame inconsistency. 

\myparagraph{Gated Self-Attention (GSA).}
To make our model adaptable to observations with different ambiguity levels, we encompass self-attention and gating mechanisms to construct a Gated Self-Attention (GSA) module. 
Here, we adopt a patch-wise self-attention block introduced in~\cite{SAN}, which efficiently computes local attention weights that vary over spatial coordinates and channels instead of sharing weights to convolve the whole feature maps like conventional CNN.
It has the form:
\begin{align}
    \mathbf{y}_i = \sum_{j \in \mathcal{R}(i)} \alpha(\mathbf{x}_{\mathcal{R}(i)})_j \odot \beta(\mathbf{x}_j), \label{eq:SA}\\ 
    \text{where} \hspace{5mm}
    \alpha(\mathbf{x}_{\mathcal{R}(i)}) = \gamma(\delta(\mathbf{x}_{\mathcal{R}(i)}))). \nonumber
\end{align}
The function $\beta$ produces the feature vectors $\beta(\mathbf{x}_j)$ that are weighted summarized by the adaptive weight vectors $\alpha(\mathbf{x}_{\mathcal{R}(i)})_j$.
The tensor $\mathbf{x}_{\mathcal{R}(i)}$ is the patch of feature vectors in a 7$\times$7 footprint $\mathcal{R}(i)$. $\alpha(\mathbf{x}_{\mathcal{R}(i)})_j$ is the attention vector at location $j$ in tensor $\alpha(\mathbf{x}_{\mathcal{R}(i)})$, corresponding spatially to the vector $x_j$ in $\mathbf{x}_{\mathcal{R}(i)}$. 
Functions $\beta$ and $\gamma$ are mappings implemented via one convolution layer, respectively. 
The function $\delta$ combines the feature vectors $x_j$ from the patch $\mathbf{x}_{\mathcal{R}(i)}$ implemented via a concatenation operation $\delta(\mathbf{x}_{\mathcal{R}(i)})) = [\phi(\mathbf{x}_i),[\psi(\mathbf{x}_j)]_{\forall j \in \mathcal{R}(i)}]$, where $\phi$ and $\psi$ are mappings implemented via one convolution layer, respectively. $\odot$ denotes the Hadamard product. 
To reduce the impact of vanishing gradients, we wrap the self-attention block in a residual structure, \new{\ie} $\mathbf{z} = \text{Conv}_r(\mathbf{y}) + \mathbf{x}$. We then equip our self-attention block with a gating mechanism to learn to control the information flows of different ambiguities,
formulated as:
\begin{align}
    \mathbf{g} = \text{sigmoid}(\text{Conv}_g(\mathbf{z})) \odot \text{tanh}(\text{Conv}_a(\mathbf{z})),
    \label{eq:gate}
\end{align}
where $\text{sigmoid}(\text{Conv}_g(\mathbf{z}))$ and $\text{tanh}(\text{Conv}_a(\mathbf{z})$ denote the gate and feature activation, respectively. 
\new{$\text{Conv}_r$, $\text{Conv}_g$, $\text{Conv}_a$ are 2D convolution layers, and the subscriptions stand for residual, gate, and attention, respectively.}

\subsection{Uncertainty mechanism}
\label{sec:uncertainty}

We design an uncertainty mechanism to not only predict the interpretable hallucination uncertainty, but also guide the learning by reducing supervision ambiguity. 
We draw inspiration from prior work where predicting the data-dependent uncertainty helped in tempering the training objective, \new{\eg} by attenuating the effect from erroneous labels in~\cite{flowUncertainty}, or by automatically balancing the loss weighting in~\cite{multitaskUncertainty}. 
To realize this idea, one may design a heuristic weighting map, like the spatially discounted reconstruction loss~\cite{yu2018generative}, but such approaches are ad-hoc and cannot adapt to different scenes automatically.
Instead, we opt to jointly learn an uncertainty map during training, which serves as a probabilistic interpretation of our model.
We build upon predictive estimation~\cite{nix1994estimating}, and infer the mean and variance of the distribution $p(O_t|I_{t-i},\mathbb{D})$, where $i=0,\dots,K$, and $\mathbb{D}$ denotes the whole dataset.
The network is trained by log-likelihood maximization (\new{\ie} negative log-likelihood minimization) and the distribution can be modelled as Laplacian (\new{\ie} corresponding to L1 loss) or Gaussian  (\new{\ie} corresponding to L2 loss) respectively.
The negative log-likelihood formulation is:
\begin{align}
    L_{1Log} = \frac{\|\mu(x)-x^*\|_1}{\sigma(x)} + \log\sigma(x),
    \label{eq:loss_u_log}
\end{align}
where $\mu(x)$ and $\sigma(x)$ are the network outputs, encoding mean and variance of the distribution.
Here, we adopt L1 loss for pixel level reconstruction and reformulate Eq.~\ref{eq:loss_u_log} integrating uncertainty $U_t$ as:
\begin{equation}
    \begin{split}
    L_{1U}^{\theta_A} = \mathbb{E}\left[ \left(\frac{\|O_t-W_t\|_1}{U_t}\right) \odot M \right. \\
    \left. + \|O_t-W_t\|_1 \odot (1- M) + U_t \vphantom{\int_1^2}\right],
    \label{eq:loss_L1u}
    \end{split}
\end{equation}
where $O_t$, $W_t$ are the predicted wide FoV RGB image and the ground truth RGB image (3-channel), respectively. $U_t$ is the estimated hallucination uncertainty map (1-channel). $M$ denotes the mask for out-of-narrow-FoV regions. Within the narrow FoV region $(1- M)$, the L1 is not weighted by uncertainty $U_t$, as new{the} present narrow FoV region has been observed.
Note that, to make the uncertainty term $U_t$ more interpretable and stabilize the training process, we constrain $U_t$ in the range (0,1) with a sigmoid function and modify the regularization term from $\log U_t$ to $U_t$ for gradient stabilization. We found that such modification also leads to better performance.
Additionally, in our recurrent framework, the previous predicted uncertainty $\{U_{t-i}\}_{i=1,\dots,j}$ are also used in future input to act as a confidence signal.

\subsection{Losses}
\label{sec:loss}

\myparagraph{Coordinate generation losses.} Following~\cite{monodepth2}, the objective in this stage is a masked photometric loss $\nu \odot L_{photo}$ and an edge-aware smoothness loss $L_{smooth}$, summarized below:
\begin{align}
    L_{CG}^{\theta_D,\theta_P} &= \nu \odot L_{photo}^{\theta_D,\theta_P} + \lambda_s L_{smooth}^{\theta_D,\theta_P},
    \label{eq:loss_CG}\\
    L_{photo}^{\theta_D,\theta_P} &=\min\limits_{t'}pe(I_t,I_{t'\rightarrow t}),  \label{eq:loss_pixel}\\ 
    pe(I_a,I_b) &= \frac{\alpha}{2}(1-\text{SSIM}(I_a,I_b)) + (1-\alpha){\|I_a-I_b\|}_1,
    \nonumber\\
    L_{smooth}^{\theta_D,\theta_P} &=|\nabla_x d^*_t|e^{-|\nabla_x I_t|} + |\nabla_y d^*_t|e^{-|\nabla_y I_t|}, 
    \label{eq:loss_smooth}
\end{align}
where $\nu=[\min\limits_{t'}pe(I_t,I_{t'\rightarrow t})]$ acts as an auto-mask for suppressing the effect of objects moving at similar speeds to the camera, and $[]$ is the Iverson bracket.  $d^*=d_t/\overline{d_t}$ is the mean-normalized inverse depth. \new{SSIM is the structural similarity~\cite{ssim}.}
We use hyper-parameters $\alpha=0.85$ and $\lambda_s=0.001$ as in~\cite{monodepth2}. 

\myparagraph{Frame aggregation losses.}
The objective of the frame aggregation stage has four loss terms: uncertainty-aware L1 reconstruction loss $L_{1U}^{\theta_A}$, perceptual reconstruction loss $L_{perc}^{\theta_A}$~\cite{lpips}, adversarial loss $L_{adv}^{\theta_A,\theta_Q}$~\cite{lsgan}, and temporal adversarial loss $L_{advT}^{\theta_A,\theta_T}$. The reconstruction losses $L_{1U}^{\theta_A}$ and $L_{perc}^{\theta_A}$ are used to regress the output towards the target ground truth, while the adversarial losses $L_{adv}^{\theta_A,\theta_Q}$ and $L_{advT}^{\theta_A,\theta_T}$ are used to encourage image photo-realism and temporal coherence. The formulations are as follows:
\begin{align}
    \begin{split}
    L_{FA}^{\theta_\mathcal{A},\theta_\mathcal{Q},\theta_\mathcal{T}} &= \lambda_1 L_{1U}^{\theta_\mathcal{A}}(O_t,W_t) + \lambda_2 L_{perc}^{\theta_\mathcal{A}}(O_t,W_t) \\
    &+ L_{adv}^{\theta_\mathcal{A},\theta_\mathcal{Q}}(O_t,W_t) + L_{advT}^{\theta_\mathcal{A},\theta_\mathcal{T}}(O_t,W_t), 
    \label{eq:loss_FA}
    \end{split}\\
    L_{perc}^{\theta_\mathcal{A}} &= \mathbb{E}\big[\|\phi(O_t)-\phi(W_t)\|_2^2\odot (1+ M))\big],
    \label{eq:loss_LPIPS}\\
   \begin{split}
    L_{adv}^{\theta_\mathcal{A},\theta_\mathcal{Q}} &= {\mathbb{E}}\big[(Q(W_t, M))^2\odot (1+ M)\big]  \\ &+{\mathbb{E}}\big[(Q(O_t, M)-1)^2\odot (1+ M)\big], 
    \label{eq:loss_GAN}
   \end{split}\\
   \begin{split}
    L_{advT}^{\theta_\mathcal{A},\theta_\mathcal{T}} &= {\mathbb{E}}\big[(\new{T}(\{W_{t-i}\}_{i=0,\dots,j}, M))^2\odot (1+ M)\big]  \\ &+{\mathbb{E}}\big[(\new{T}(\{O_{t-i}\}_{i=0,\dots,j}, M)-1)^2\odot (1+ M)\big], 
    \label{eq:loss_GAN_temp}
   \end{split}
\end{align}
where $\lambda_1=3$, $\lambda_2=10$. $\phi$ is a VGG network~\cite{VGG}. 

\section{Experiments}
\label{sec:exp}

To evaluate FoV-Net w.r.t. its FoV extrapolation capabilities, we provide both qualitative and quantitative results. 
For more video results and training/implementation details, please visit the supplementary materials.

\myparagraph{Dataset.}
Our method is evaluated on two widely used datasets: raw KITTI sequences~\cite{kitti} using the split from Eigen~\etal\cite{eigen2014depth}, and Cityscapes sequences~\cite{Cordts2016Cityscapes}. 
To reduce the redundant information in videos, we downsample the frame rate to 1/2 and 1/3 for KITTI and Cityscapes sequences.
Therefore, we have 39350/4382 and 59526/14992 train/val frames on KITTI and Cityscapes, respectively.
During testing, we prepare each video sample with 10 and 5 target frames for KITTI and Cityscapes.
Totally, there are 635 (6350 frames) and 1525 test video samples (7625 frames) in KITTI and Cityscapes, respectively. 
We use both monocular left and right camera sequences for training and validation, but only use left camera data for testing. 
In each forward pass, we use 6 successive frames ($k=5$ past + 1 present) as input.
The narrow FoV ratio is set to 0.5 in our experiments, \new{\ie} all 6 frames are cropped 25\% on both left and right sides to mimic the narrow FoV.

\begin{figure*}
\centering
\begin{subfigure}{.12\linewidth}
\centering
\renewcommand{\arraystretch}{2.6} 
\setlength\tabcolsep{1pt}
\begin{tabular}{c}
Inputs \\
Mono~\cite{monodepth2} \\
VF~\cite{liu2017voxelflow} \\
LGTSM~\cite{LGTSM} \\
\makecell{Mono-LGTSM} \\
Ours\\
\makecell{Uncertainty}\\
\makecell{Target}\\
\end{tabular}
\end{subfigure}
%
\begin{subfigure}{.41\linewidth}
\raisebox{-\totalheight}{
    \includegraphics[height=9.cm]{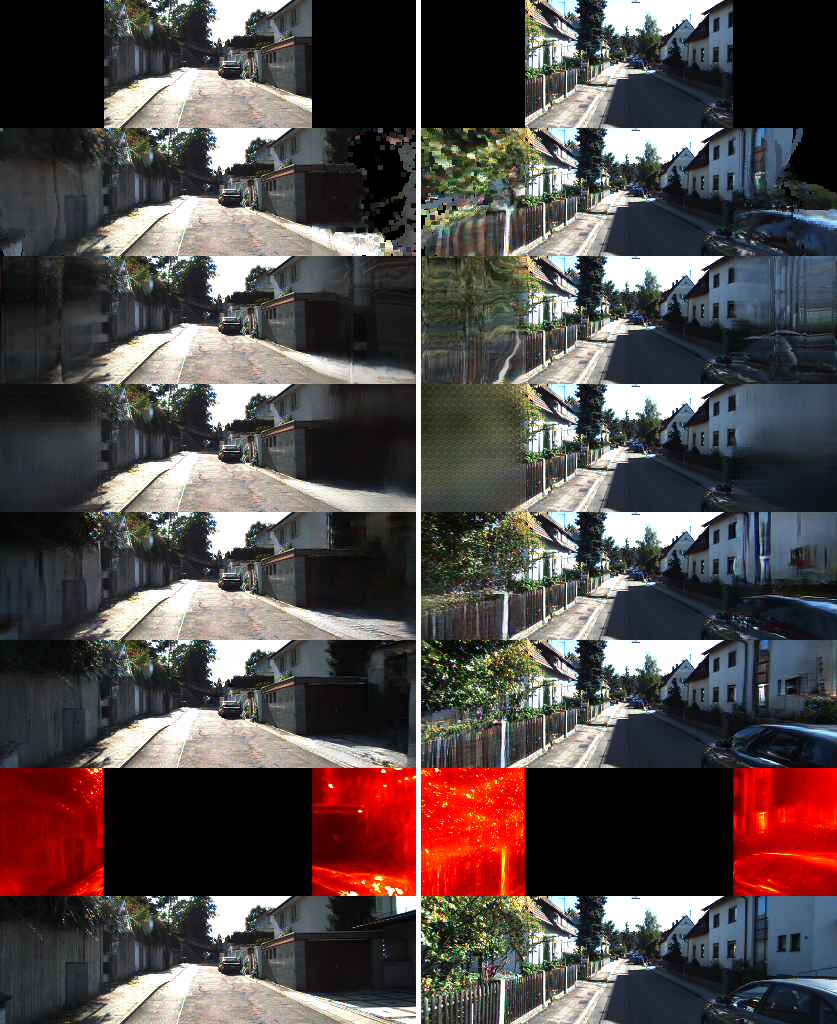}\\
}
\end{subfigure}
\hspace{7mm}
\begin{subfigure}{.41\linewidth}
\raisebox{-\totalheight}{
    \includegraphics[height=9.cm]{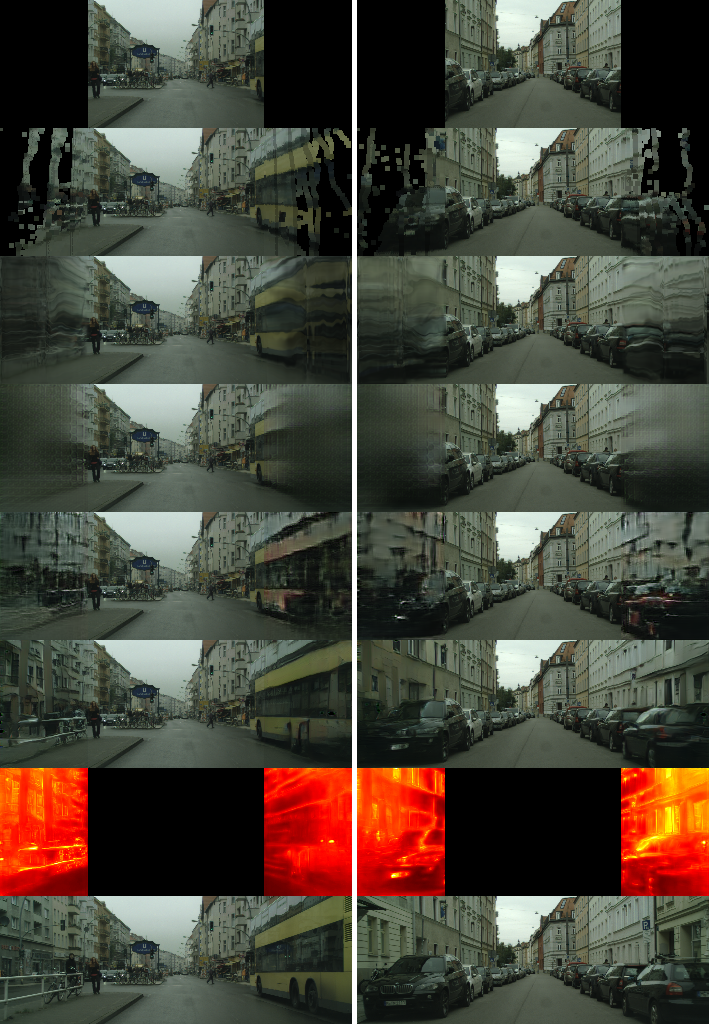}\\
}
\end{subfigure}
\caption{The qualitative comparisons on KITTI (column 1-2) and Cityscapes (column 3-4). \new{The video version is included in our supplementary PDF.}}
\label{fig:kitti_cs_baselines}
\end{figure*}

\myparagraph{Metrics.}
For quantitative evaluation,
four image quality metrics are used: Structural Similarity (SSIM)~\cite{ssim}, Learned Perceptual Image Patch Similarity (LPIPS)~\cite{lpips}, \new{Fréchet} Inception Distance (FID)~\cite{fid}, and Fréchet Video Distance (FVD)~\cite{fvd}. SSIM and LPIPS are used to evaluate the similarity between the result and the ground truth. FID and FVD are used for realistic appearance evaluation on the image level and video level, respectively. Thus, FVD can reflect both appearance realism and temporal coherence of the results.
For SSIM, higher scores are better. 
For LPIPS, FID, and FVD, lower scores are better. 
We use VGG~\cite{VGG} pre-trained on ImageNet as the feature extractor of LPIPS.
To evaluate how significant the modeled hallucination uncertainties are, we use sparsification plots and Area Under the Sparsification Error (AUSE, the lower, the better) which quantify how close the estimate is to the oracle uncertainty, as in~\cite{depthUncertainty}. More details are given in ``Uncertainty results''.

\begin{table*}[t]
    \begin{minipage}{0.65\textwidth}
    \centering
    \setlength{\tabcolsep}{5.5pt} 
    \caption{The quantitative results on KITTI and Cityscapes. 
    }
    \begin{tabular}
    {@{\extracolsep{\fill}} l c c c c | c c c c}
    \toprule 
    & \multicolumn{4}{c}{KITTI} & \multicolumn{4}{c}{Cityscapes} \\
    \cmidrule{2-5} \cmidrule{6-9}
    Model & SSIM$\uparrow$ & LPIPS$\downarrow$ & FID$\downarrow$ & FVD$\downarrow$ & SSIM$\uparrow$ & LPIPS$\downarrow$ & FID$\downarrow$ & FVD$\downarrow$ \\
    \midrule[0.6pt]	
        Mono~\cite{monodepth2} & 0.6803 & 0.2975 & 31.14 & 269.0  & 0.6444 & 0.3375 & 44.90 & 552.8 \\
        VF~\cite{liu2017voxelflow} & 0.7184 & 0.3049 & 33.51 & 360.8  & 0.7844 & 0.2936 & 38.78 & 491.5 \\
        LGTSM~\cite{LGTSM} & \textbf{0.7369} & 0.2905 & 52.57 & 495.7  & \textbf{0.7959} & 0.2845 & 72.86 & 572.9 \\
        Mono-LGTSM & 0.7028 & 0.2798 & 18.95 & 201.6  & 0.7752 & 0.2866 & 51.18 & 457.8 \\
    \midrule[0.6pt]	
        Ours & 0.7162 & \textbf{0.2294} & \textbf{10.94} & \textbf{82.71}  & 0.7539 & \textbf{0.2220} & \textbf{9.27} & \textbf{203.4} \\
    \bottomrule[1pt]
    \end{tabular}
    \label{tab:kitti_cs}
    \end{minipage}
    \begin{minipage}{0.35\textwidth}
    \centering
    \setlength{\tabcolsep}{5.5pt} 
    \caption{Ablation study results on KITTI.} 
    \begin{tabular}
    {@{\extracolsep{\fill}} l c c c c c}
    \toprule 
    Model & SSIM$\uparrow$ & LPIPS$\downarrow$ & FID$\downarrow$ & FVD$\downarrow$ \\
    \midrule[0.6pt]	
        Base & 0.7020 & 0.2407 & 12.26 & 99.75 \\
        w/o AFA & 0.7016 & 0.2377 & 11.25 & 91.92 \\
        w/o GSA & 0.7160 & 0.2386 & 11.94 & 98.67 \\
        w/o U & \textbf{0.7200} & 0.2312 & 11.37 & 89.70 \\
    \midrule[0.6pt]	
        w/o 3D & 0.6830 & 0.2618 & 16.71 & 171.1 \\
        w/o Recur & 0.7145 & 0.2331 & 11.11 & 109.5 \\
    \midrule[0.6pt]	
        Ours & 0.7162 & \textbf{0.2294} & \textbf{10.94} & \textbf{82.71} \\
    \bottomrule[1pt]
    \end{tabular}
    \label{tab:ablation}
    \end{minipage}
\end{table*}

\subsection{Comparisons}
\label{sec:comparisons}

\myparagraph{Baselines.}
As no prior work is directly comparable to our setting, we compare against geometric method Mono~\cite{monodepth2}, flow-based video prediction method VoxelFlow~\cite{liu2017voxelflow}, and video completion method LGTSM~\cite{LGTSM}, the closest alternatives. We also evaluate the Mono+LGTSM (Mono-LGTSM) setting, where pixels of past frames are first propagated by Mono before fed into LGTSM.

\myparagraph{Hallucination Results.}
Qualitative comparisons with other alternatives are provided in Fig.~\ref{fig:kitti_cs_baselines}. We observe that our method can synthesize more realistic and perceptually appealing results. For example, our method can produce hallucinations with less distortion in the 2nd column result, and better preserve the object appearance with less artifacts in the 3rd and 4th column results. This trend is also reflected from the quantitative measurements (SSIM, LPIPS, and FID scores) in Tab.~\ref{tab:kitti_cs}. Our method generates sharper and more photo-realistic results than VF~\cite{liu2017voxelflow} and LGTSM~\cite{LGTSM}, but the latter have a higher SSIM, probably due to blur, something also argued in~\cite{johnson2016perceptual,Shi-CVPR16-Superres,ma2017pose}.
\new{While LPIPS is generally more consistent with human perception and has been wildly adopted in recent image synthesis works ~\cite{ren2020deep,LGTSM}}
Furthermore, our method alleviates the flickering via the simple recurrent strategy which can retain temporal consistency. 
Note that, the hallucination of unobserved regions is improved as more surrounding observations become available in later frames.
The FVD score is also consistent with such observations.
In addition to blur and distortion, Mono~\cite{monodepth2} and VF~\cite{liu2017voxelflow} suffer from missing pixels. LGTSM~\cite{LGTSM} cannot propagate the information to the out-of-view regions correctly and thus performs poorly. When using Mono and LGTSM together, we can combine their merits and arrive at better results. This indicates that both 3D cues and hallucination capability are key to the FoV extrapolation problem. FoV-Net gets the best of both worlds, and further addresses ambiguity, leading to the best outcome. 

\begin{figure*}[t]
    \centering
    \footnotesize
    \includegraphics[width=1\linewidth]{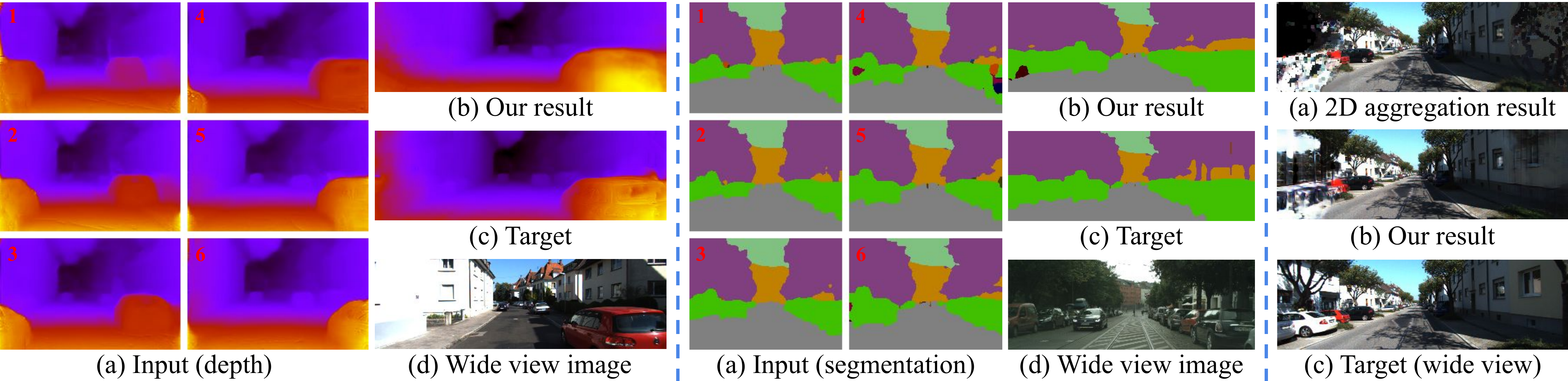}\\
    \caption{Left: depth extrapolation. Middle: semantic segmentation extrapolation. Right: failure case. Zoom in for more details.}
  \label{fig:depth_seg_failure}
\end{figure*}

\begin{figure}[t!]
    \begin{subfigure}[t]{{0.49\columnwidth}}
    \centering 
    \includegraphics[height=0.8 in]{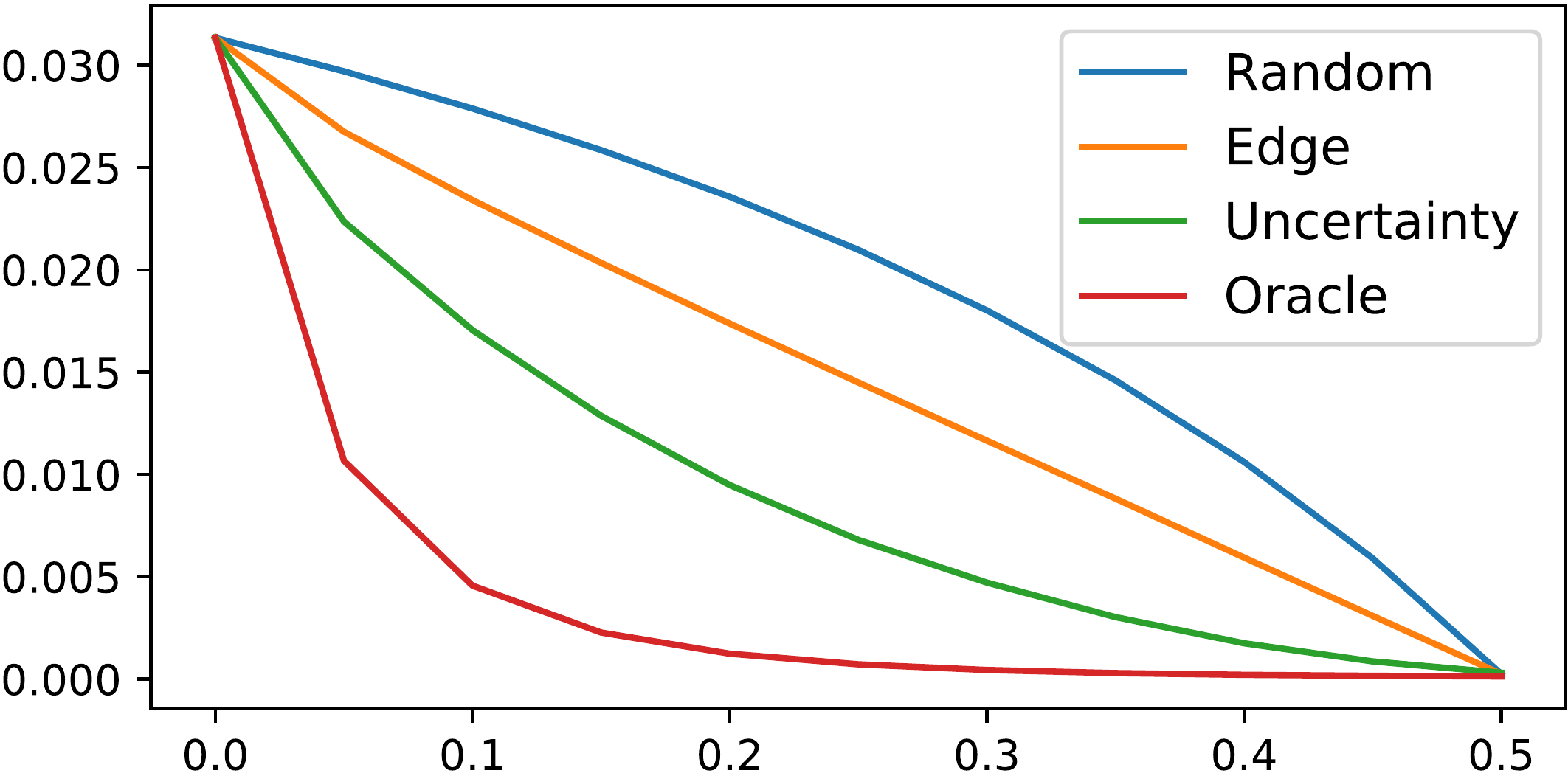}
    \caption{KITTI}
    \end{subfigure}
    \begin{subfigure}[t]{{0.49\columnwidth}}
    \centering 
    \includegraphics[height=0.8 in]{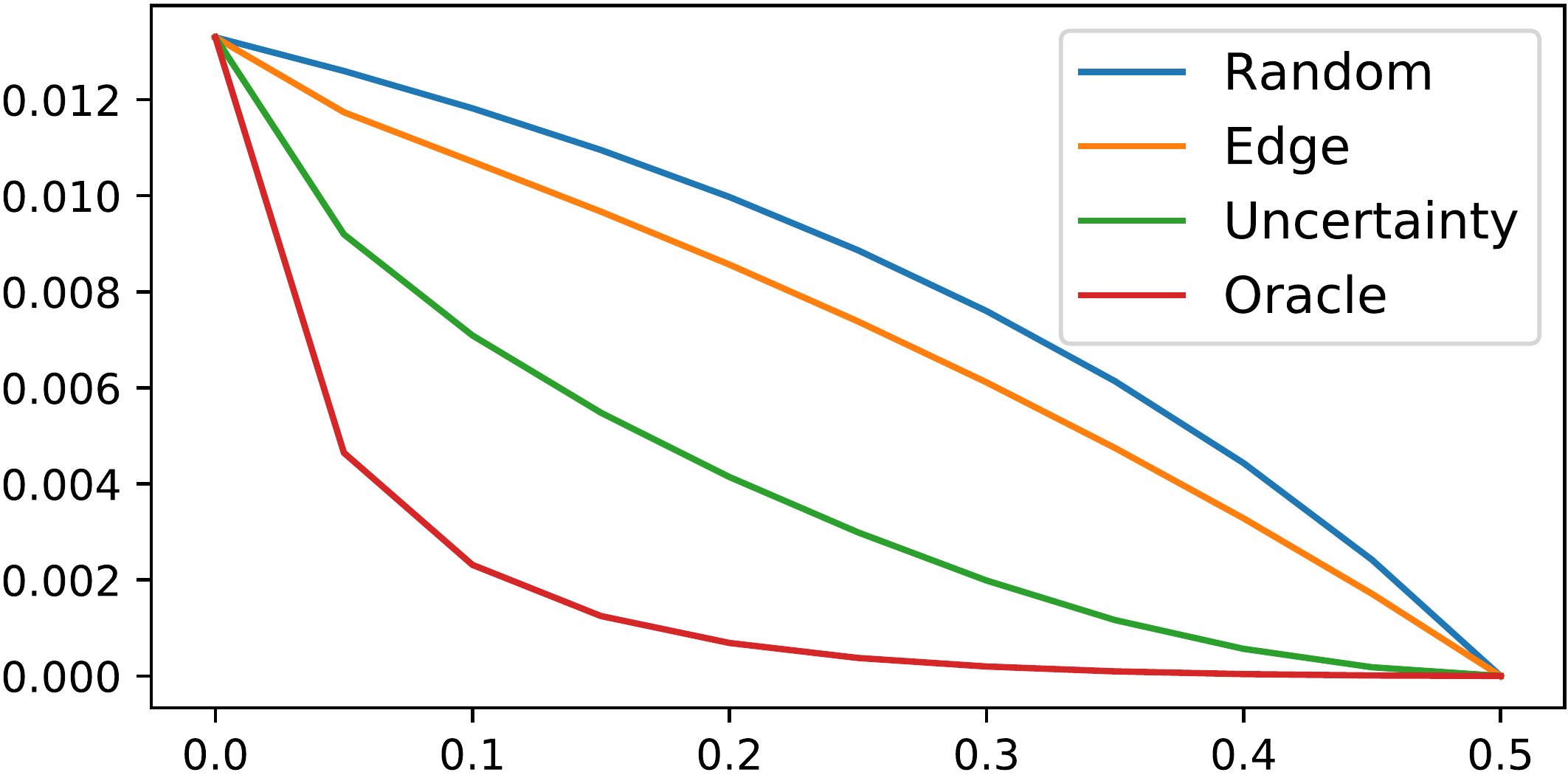}
    \caption{Cityscapes}
    \end{subfigure}
    \caption{\new{The sparsification plots. The x-axis denotes the fraction of removed pixels, and the y-axis shows the MSE on the remaining pixels. MSE converges to 0 after removing all pixels in the out-of-view regions. Zoom in for more details.}}
  \label{fig:uncertainty}
\end{figure}

\myparagraph{Uncertainty results.}
To evaluate the significance of estimated uncertainty, we adopt a pixel-wise metric Mean-Square-Error (MSE) to sort all pixels in each hallucinated wide FoV image in order of descending uncertainty. Then, we iteratively remove a subset of pixels in the out-of-view regions
(\new{\ie} 5\% in our experiments) and compute MSE on the remaining to plot a curve that is supposed to shrink if the uncertainty properly encodes the \new{hallucinated image's errors} (see Fig.~\ref{fig:uncertainty}). 
An ideal sparsification (oracle) is obtained by sorting pixels in descending order of the MSE magnitude. In contrast, a random uncertainty is to remove the pixels randomly each time. \new{Besides, we observe that there is usually high uncertainty in the edge part. Therefore, we also construct an edge uncertainty baseline which adopts an estimated soft edge map~\cite{hed_edge} to approximate the hallucination uncertainty.}
As we use 0.5 as our narrow view ratio, the curves decrease to zero when 50\% of
pixels are extracted. The AUSE scores on KITTI are 0.0049, 0.0105, \new{0.0149}, and 0.0197 for oracle, uncertainty, \new{edge}, and random settings, respectively. The AUSE scores on Cityscapes are 0.0021, 0.0042, \new{0.0070}, and 0.0080 for oracle, uncertainty, \new{edge}, and random settings, respectively.
As shown in Fig.~\ref{fig:uncertainty}, our method can successfully estimate the hallucination uncertainty which reasonably indicates the prediction errors. For example, in the 2nd column result of Fig.~\ref{fig:kitti_cs_baselines}, the uncertainty is high on the thin pole and car edges, as the depth estimation \new{is} noisy there.
In the 4th column result of Fig.~\ref{fig:kitti_cs_baselines}, the uncertainty is  high on the car edges and unobserved regions (black holes in the Mono result). 
\myparagraph{Ablation study}
is performed on KITTI for each component of FoV-Net. The ablative settings are as follows.
\textbf{Base:} removing the AFA module, GSA module, and the uncertainty mechanism.
\textbf{w/o AFA:} using temporal average pooling to replace the AFA module.
\textbf{w/o GSA:} removing the GSA modules.
\textbf{w/o U:} removing the uncertainty mechanism.
\textbf{w/o 3D:} removing the coordinate generation stage and the feature propagation operation in the frame aggregation network.
\textbf{w/o Recur:} removing temporal modeling, \new{\ie} no recurrent feed-forward and temporal discriminator $\mathcal{T}$.
The quantitative results are in Tab.~\ref{tab:ablation}. When equipped with our AFA and GSA modules, the results are perceptually closer to the wide FoV targets (in terms of LPIPS score decreasing), as well as more photo-realistic (in terms of FID/FVD scores decreasing) and temporal consistency (in terms of FVD score decreasing). Besides, our interpretable uncertainty mechanism could improve the performance moderately, indicating its ability to guide the learning by reducing supervision ambiguity. In addition, the 3D cues and temporal modeling are also important to good performance. 

\subsection{Extended Applications}
\label{sec:extensions}
In this section, we provide several extensions (also see supplementary material).
We further apply our method to depth extrapolation (Fig.~\ref{fig:depth_seg_failure} left) and semantic extrapolation (Fig.~\ref{fig:depth_seg_failure} middle). Results show that our method can benefit other vision perception tasks as well, which are important for robotic motion planning and navigation. 

\myparagraph{Depth extrapolation.}
To extrapolate the wide FoV depth, the input will be the narrow FoV depth maps. Our frame aggregation network is then trained with the scale-invariant depth regression loss~\cite{eigen2014depth}.
Both the input and target depth maps are estimated by our depth network $D$. 

\myparagraph{Semantic segmentation extrapolation.}
To infer the wide FoV semantic segmentation, the input is the narrow FoV segmentation maps, and our frame aggregation network is then trained with a cross-entropy loss.
Both the input and target segmentation maps are estimated by HRNet~\cite{HRNet}.

\section{Limitations and conclusions}
\new{
While FoV-Net has achieved good results, there remain a plethora of avenues for future work.
First, our FoV-Net may be not robust to fast moving objects.
For example, in Fig.~\ref{fig:depth_seg_failure} right, the blurred left region is due to the white moving car which causes serious propagation errors.
One potential solution is to extend the FoV extrapolation in the 3D space with a multi-sensor system and model these moving objects explicitly.}
\new{Second, in this work, we focus on extrapolating the present frame to a limited wide view. While 360\degree FoV extrapolation and future prediction could be more helpful in some cases. We plan to extend our method both spatially and temporally to enable 360\degree FoV future prediction.
In conclusion, we present FoV-Net to tackle the \textit{FoV extrapolation} problem. Our framework propagates and aggregates the information observed from the past and current narrow FoV frames to generate the current wide FoV, as well as predicts the hallucination uncertainty. We take a significant step to endow machines with hallucination ability, and we believe such an ability can benefit the robotics community.}

\printbibliography

@inproceedings{kitti,
  title={Are we ready for autonomous driving? the kitti vision benchmark suite},
  author={Geiger, Andreas and Lenz, Philip and Urtasun, Raquel},
  booktitle={CVPR},
  year={2012},
}

@inproceedings{eigen2014depth,
  title={Depth map prediction from a single image using a multi-scale deep network},
  author={Eigen, David and Puhrsch, Christian and Fergus, Rob},
  booktitle={NeurIPS},
  year={2014}
}

@inproceedings{bansal2018recycle,
  title={Recycle-gan: Unsupervised video retargeting},
  author={Bansal, Aayush and Ma, Shugao and Ramanan, Deva and Sheikh, Yaser},
  booktitle={ECCV},
  year={2018}
}

@inproceedings{wang2018video,
  title={Video-to-video synthesis},
  author={Wang, Ting-Chun and Liu, Ming-Yu and Zhu, Jun-Yan and Liu, Guilin and Tao, Andrew and Kautz, Jan and Catanzaro, Bryan},
  booktitle={NeurIPS},
  year={2018}
}

@inproceedings{chen2017coherent,
  title={Coherent online video style transfer},
  author={Chen, Dongdong and Liao, Jing and Yuan, Lu and Yu, Nenghai and Hua, Gang},
  booktitle={ICCV},
  year={2017}
}

@inproceedings{shi2016real,
  title={Real-time single image and video super-resolution using an efficient sub-pixel convolutional neural network},
  author={Shi, Wenzhe and Caballero, Jose and Husz{\'a}r, Ferenc and Totz, Johannes and Aitken, Andrew P and Bishop, Rob and Rueckert, Daniel and Wang, Zehan},
  booktitle={CVPR},
  year={2016}
}

@inproceedings{ren2020deep,
  title={Deep image spatial transformation for person image generation},
  author={Ren, Yurui and Yu, Xiaoming and Chen, Junming and Li, Thomas H and Li, Ge},
  booktitle={CVPR},
  year={2020}
}

@inproceedings{liu2017voxelflow,
  author={Liu, Ziwei and Yeh, Raymond A and Tang, Xiaoou and Liu, Yiming and Agarwala, Aseem},
 title = {Video Frame Synthesis using Deep Voxel Flow},
 booktitle = {ICCV},
 year = {2017} 
}

@inproceedings{denton2018stochastic,
  title={Stochastic video generation with a learned prior},
  author={Denton, Emily and Fergus, Rob},
  booktitle={ICML},
  year={2018}
}

@inproceedings{byeon2018contextvp,
  title={Contextvp: Fully context-aware video prediction},
  author={Byeon, Wonmin and Wang, Qin and Kumar Srivastava, Rupesh and Koumoutsakos, Petros},
  booktitle={ECCV},
  year={2018}
}

@article{lotter2016deep,
  title={Deep predictive coding networks for video prediction and unsupervised learning},
  author={Lotter, William and Kreiman, Gabriel and Cox, David},
  journal={arXiv preprint arXiv:1605.08104},
  year={2016}
}

@inproceedings{gao2019disentangling,
  title={Disentangling propagation and generation for video prediction},
  author={Gao, Hang and Xu, Huazhe and Cai, Qi-Zhi and Wang, Ruth and Yu, Fisher and Darrell, Trevor},
  booktitle={ICCV},
  year={2019}
}

@inproceedings{liu2018dyan,
  title={Dyan: A dynamical atoms-based network for video prediction},
  author={Liu, Wenqian and Sharma, Abhishek and Camps, Octavia and Sznaier, Mario},
  booktitle={ECCV},
  year={2018}
}

@article{fvd,
  title={Towards Accurate Generative Models of Video: A New Metric \& Challenges},
  author={Unterthiner, Thomas and van Steenkiste, Sjoerd and Kurach, Karol and Marinier, Raphael and Michalski, Marcin and Gelly, Sylvain},
  journal={arXiv preprint arXiv:1812.01717},
  year={2018}
}

@article{ssim,
  title={Image quality assessment: from error visibility to structural similarity},
  author={Wang, Zhou and Bovik, Alan C and Sheikh, Hamid R and Simoncelli, Eero P and others},
  journal={IEEE transactions on image processing},
  volume={13},
  number={4},
  pages={600--612},
  year={2004}
}

@inproceedings{lpips,
  title={The Unreasonable Effectiveness of Deep Features as a Perceptual Metric},
  author={Zhang, Richard and Isola, Phillip and Efros, Alexei A and Shechtman, Eli and Wang, Oliver},
  booktitle={CVPR},
  year={2018}
}

@inproceedings{fid,
  title={Gans trained by a two time-scale update rule converge to a local nash equilibrium},
  author={Heusel, Martin and Ramsauer, Hubert and Unterthiner, Thomas and Nessler, Bernhard and Hochreiter, Sepp},
  booktitle={NeurIPS},
  year={2017}
}

@inproceedings{jaderberg2015spatial,
  title={Spatial transformer networks},
  author={Jaderberg, Max and Simonyan, Karen and Zisserman, Andrew},
  booktitle={NeurIPS},
  year={2015}
}

@inproceedings{monodepth2,
  title={Digging into self-supervised monocular depth estimation},
  author={Godard, Cl{\'e}ment and Mac Aodha, Oisin and Firman, Michael and Brostow, Gabriel J},
  booktitle={ICCV},
  year={2019}
}

@inproceedings{HRNet,
  title={Deep high-resolution representation learning for human pose estimation},
  author={Sun, Ke and Xiao, Bin and Liu, Dong and Wang, Jingdong},
  booktitle={CVPR},
  year={2019}
}

@inproceedings{park2017transformation,
  title={Transformation-grounded image generation network for novel 3d view synthesis},
  author={Park, Eunbyung and Yang, Jimei and Yumer, Ersin and Ceylan, Duygu and Berg, Alexander C},
  booktitle={CVPR},
  year={2017}
}

@inproceedings{ma2017pose,
  title={Pose guided person image generation},
  author={Ma, Liqian and Jia, Xu and Sun, Qianru and Schiele, Bernt and Tuytelaars, Tinne and Van Gool, Luc},
  booktitle={NeurIPS},
  year={2017}
}

@inproceedings{sun2018multi,
  title={Multi-view to Novel view: Synthesizing novel views with Self-Learned Confidence},
  author={Sun, Shao-Hua and Huh, Minyoung and Liao, Yuan-Hong and Zhang, Ning and Lim, Joseph J},
  booktitle={ECCV},
  year={2018}
}

@article{chen2019nvs,
  title={NVS Machines: Learning Novel View Synthesis with Fine-grained View Control},
  author={Chen, Xu and Song, Jie and Hilliges, Otmar},
  journal={arXiv preprint arXiv:1901.01880},
  year={2019}
}

@inproceedings{choi2019extreme,
  title={Extreme view synthesis},
  author={Choi, Inchang and Gallo, Orazio and Troccoli, Alejandro and Kim, Min H and Kautz, Jan},
  booktitle={ICCV},
  year={2019}
}

@article{hedman2018deep,
  title={Deep blending for free-viewpoint image-based rendering},
  author={Hedman, Peter and Philip, Julien and Price, True and Frahm, Jan-Michael and Drettakis, George and Brostow, Gabriel},
  journal={ACM TOG},
  volume={37},
  number={6},
  pages={1--15},
  year={2018},
}

@inproceedings{flynn2019deepview,
  title={DeepView: View synthesis with learned gradient descent},
  author={Flynn, John and Broxton, Michael and Debevec, Paul and DuVall, Matthew and Fyffe, Graham and Overbeck, Ryan and Snavely, Noah and Tucker, Richard},
  booktitle={CVPR},
  year={2019}
}

@inproceedings{deepfillv2,
  title={Free-form image inpainting with gated convolution},
  author={Yu, Jiahui and Lin, Zhe and Yang, Jimei and Shen, Xiaohui and Lu, Xin and Huang, Thomas S},
  booktitle={ICCV},
  year={2019}
}

@inproceedings{yu2018generative,
  title={Generative image inpainting with contextual attention},
  author={Yu, Jiahui and Lin, Zhe and Yang, Jimei and Shen, Xiaohui and Lu, Xin and Huang, Thomas S},
  booktitle={CVPR},
  year={2018}
}

@article{huang2016temporally,
  title={Temporally coherent completion of dynamic video},
  author={Huang, Jia-Bin and Kang, Sing Bing and Ahuja, Narendra and Kopf, Johannes},
  journal={ACM TOG},
  volume={35},
  number={6},
  pages={1--11},
  year={2016},
}

@inproceedings{LGTSM,
  title={Learnable gated temporal shift module for deep video inpainting},
  author={Chang, Ya-Liang and Liu, Zhe Yu and Lee, Kuan-Ying and Hsu, Winston},
  booktitle={BMVC},
  year={2019}
}

@inproceedings{chang2019free,
  title={Free-form video inpainting with 3d gated convolution and temporal patchgan},
  author={Chang, Ya-Liang and Liu, Zhe Yu and Lee, Kuan-Ying and Hsu, Winston},
  booktitle={ICCV},
  year={2019}
}

@inproceedings{kim2019deep,
  title={Deep video inpainting},
  author={Kim, Dahun and Woo, Sanghyun and Lee, Joon-Young and Kweon, In So},
  booktitle={CVPR},
  year={2019}
}

@inproceedings{wang2019video,
  title={Video inpainting by jointly learning temporal structure and spatial details},
  author={Wang, Chuan and Huang, Haibin and Han, Xiaoguang and Wang, Jue},
  booktitle={AAAI},
  year={2019}
}

@inproceedings{Gao-ECCV-FGVC,
    author    = {Gao, Chen and Saraf, Ayush and Huang, Jia-Bin and Kopf, Johannes},
    title     = {Flow-edge Guided Video Completion},
    booktitle = {ECCV},
    year      = {2020}
}

@inproceedings{icra2020attentive,
  title={Attentive Task-Net: Self Supervised Task-Attention Network for Imitation Learning using Video Demonstration},
  author={Ramachandruni, Kartik and Babu, Madhu and Majumder, Anima and Dutta, Samrat and Kumar, Swagat},
  booktitle={IEEE Int. Conf. on Robotics and Automation (ICRA)},
  year={2020},
}

@inproceedings{zhao2018paragraph,
  title={Paragraph-level neural question generation with maxout pointer and gated self-attention networks},
  author={Zhao, Yao and Ni, Xiaochuan and Ding, Yuanyuan and Ke, Qifa},
  booktitle={Conference on Empirical Methods in Natural Language Processing (EMNLP)},
  year={2018}
}

@inproceedings{lai2019gated,
  title={A Gated Self-attention Memory Network for Answer Selection},
  author={Lai, Tuan and Tran, Quan Hung and Bui, Trung and Kihara, Daisuke},
  booktitle={Conference on Empirical Methods in Natural Language Processing (EMNLP)},
  year={2019}
}

@inproceedings{dhingra2016gated,
  title={Gated-attention readers for text comprehension},
  author={Dhingra, Bhuwan and Liu, Hanxiao and Yang, Zhilin and Cohen, William W and Salakhutdinov, Ruslan},
  booktitle={ACL},
  year={2017}
}

@inproceedings{tran2017generative,
  title={A generative attentional neural network model for dialogue act classification},
  author={Tran, Quan Hung and Haffari, Gholamreza and Zukerman, Ingrid},
  booktitle={ACL},
  year={2017}
}

@inproceedings{SAGAN,
  title={Self-attention generative adversarial networks},
  author={Zhang, Han and Goodfellow, Ian and Metaxas, Dimitris and Odena, Augustus},
  booktitle={ICML},
  year={2019}
}

@inproceedings{brock2018large,
  title={Large scale gan training for high fidelity natural image synthesis},
  author={Brock, Andrew and Donahue, Jeff and Simonyan, Karen},
  booktitle={ICLR},
  year={2019}
}

@inproceedings{parmar2018image,
  title={Image transformer},
  author={Parmar, Niki and Vaswani, Ashish and Uszkoreit, Jakob and Kaiser, {\L}ukasz and Shazeer, Noam and Ku, Alexander and Tran, Dustin},
  booktitle={ICML},
  year={2018}
}

@inproceedings{jia2016dynamic,
  title={Dynamic filter networks},
  author={Jia, Xu and De Brabandere, Bert and Tuytelaars, Tinne and Gool, Luc V},
  booktitle={NeurIPS},
  year={2016}
}

@inproceedings{wang2018non,
  title={Non-local neural networks},
  author={Wang, Xiaolong and Girshick, Ross and Gupta, Abhinav and He, Kaiming},
  booktitle={CVPR},
  year={2018}
}

@inproceedings{hu2019local,
  title={Local relation networks for image recognition},
  author={Hu, Han and Zhang, Zheng and Xie, Zhenda and Lin, Stephen},
  booktitle={ICCV},
  year={2019}
}

@inproceedings{SAN,
  title={Exploring Self-attention for Image Recognition},
  author={Zhao, Hengshuang and Jia, Jiaya and Koltun, Vladlen},
  booktitle={CVPR},
  year={2020}
}

@inproceedings{nix1994estimating,
  title={Estimating the mean and variance of the target probability distribution},
  author={Nix, David A and Weigend, Andreas S},
  booktitle={ICNN},
  year={1994},
}

@inproceedings{depthUncertainty,
  title={On the uncertainty of self-supervised monocular depth estimation},
  author={Poggi, Matteo and Aleotti, Filippo and Tosi, Fabio and Mattoccia, Stefano},
  booktitle={CVPR},
  year={2020}
}

@inproceedings{multitaskUncertainty,
  title={Multi-task learning using uncertainty to weigh losses for scene geometry and semantics},
  author={Kendall, Alex and Gal, Yarin and Cipolla, Roberto},
  booktitle={CVPR},
  year={2018}
}

@inproceedings{he2019bounding,
  title={Bounding box regression with uncertainty for accurate object detection},
  author={He, Yihui and Zhu, Chenchen and Wang, Jianren and Savvides, Marios and Zhang, Xiangyu},
  booktitle={CVPR},
  year={2019}
}

@inproceedings{flowUncertainty,
  title={Uncertainty estimates and multi-hypotheses networks for optical flow},
  author={Ilg, Eddy and Cicek, Ozgun and Galesso, Silvio and Klein, Aaron and Makansi, Osama and Hutter, Frank and Brox, Thomas},
  booktitle={ECCV},
  year={2018}
}

@article{yang2020d3vo,
  title={D3VO: Deep Depth, Deep Pose and Deep Uncertainty for Monocular Visual Odometry},
  author={Yang, Nan and von Stumberg, Lukas and Wang, Rui and Cremers, Daniel},
  journal={arXiv preprint arXiv:2003.01060},
  year={2020}
}

@inproceedings{huang2018efficient,
  title={Efficient uncertainty estimation for semantic segmentation in videos},
  author={Huang, Po-Yu and Hsu, Wan-Ting and Chiu, Chun-Yueh and Wu, Ting-Fan and Sun, Min},
  booktitle={ECCV},
  year={2018}
}

@article{loquercio2020general,
  title={A general framework for uncertainty estimation in deep learning},
  author={Loquercio, Antonio and Segu, Mattia and Scaramuzza, Davide},
  journal={IEEE Robotics and Automation Letters (RAL)},
  volume={5},
  number={2},
  pages={3153--3160},
  year={2020},
  publisher={IEEE}
}

@inproceedings{hed_edge,
  title={Holistically-nested edge detection},
  author={Xie, Saining and Tu, Zhuowen},
  booktitle={ICCV},
  year={2015}
}

@article{lee2019video,
  title={Video Extrapolation Using Neighboring Frames},
  author={Lee, Sangwoo and Lee, Jungjin and Kim, Bumki and Kim, Kyehyun and Noh, Junyong},
  journal={ACM TOG},
  volume={38},
  number={3},
  pages={1--13},
  year={2019},
}

@article{choi2020deep,
  title={Deep Iterative Frame Interpolation for Full-frame Video Stabilization},
  author={Choi, Jinsoo and Kweon, In So},
  journal={ACM TOG},
  volume={39},
  number={1},
  pages={1--9},
  year={2020},
}

@inproceedings{zhang2014parallax,
  title={Parallax-tolerant image stitching},
  author={Zhang, Fan and Liu, Feng},
  booktitle={CVPR},
  year={2014}
}

@article{guo2016joint,
  title={Joint video stitching and stabilization from moving cameras},
  author={Guo, Heng and Liu, Shuaicheng and He, Tong and Zhu, Shuyuan and Zeng, Bing and Gabbouj, Moncef},
  journal={IEEE transactions on image processing (TIP)},
  volume={25},
  number={11},
  pages={5491--5503},
  year={2016},
  publisher={IEEE}
}

@inproceedings{Cordts2016Cityscapes,
title={The Cityscapes Dataset for Semantic Urban Scene Understanding},
author={Cordts, Marius and Omran, Mohamed and Ramos, Sebastian and Rehfeld, Timo and Enzweiler, Markus and Benenson, Rodrigo and Franke, Uwe and Roth, Stefan and Schiele, Bernt},
booktitle={CVPR},
year={2016}
}

@inproceedings{lsgan,
  title={Least squares generative adversarial networks},
  author={Mao, Xudong and Li, Qing and Xie, Haoran and Lau, Raymond YK and Wang, Zhen and Paul Smolley, Stephen},
  booktitle={ICCV},
  year={2017}
}

@inproceedings{johnson2016perceptual,
  title={Perceptual losses for real-time style transfer and super-resolution},
  author={Johnson, Justin and Alahi, Alexandre and Fei-Fei, Li},
  booktitle={ECCV},
  year={2016}
}

@inproceedings{VGG,
  title={Very deep convolutional networks for large-scale image recognition},
  author={Simonyan, Karen and Zisserman, Andrew},
  booktitle={ICLR},
  year={2015}
}

@inproceedings{Shi-CVPR16-Superres,
  author    = {Wenzhe Shi and
               Jose Caballero and
               Ferenc Huszar and
               Johannes Totz and
               Andrew P. Aitken and
               Rob Bishop and
               Daniel Rueckert and
               Zehan Wang},
  title     = {Real-Time Single Image and Video Super-Resolution Using an Efficient Sub-Pixel Convolutional Neural Network},
  booktitle = {CVPR},
  year      = {2016}
}

@inproceedings{wang2019reinforced,
  title={Reinforced cross-modal matching and self-supervised imitation learning for vision-language navigation},
  author={Wang, Xin and Huang, Qiuyuan and Celikyilmaz, Asli and Gao, Jianfeng and Shen, Dinghan and Wang, Yuan-Fang and Wang, William Yang and Zhang, Lei},
  booktitle={CVPR},
  year={2019}
}

@inproceedings{jayaraman2018learning,
  title={Learning to look around: Intelligently exploring unseen environments for unknown tasks},
  author={Jayaraman, Dinesh and Grauman, Kristen},
  booktitle={CVPR},
  year={2018}
}

@article{makhataeva2020augmented,
  title={Augmented Reality for Robotics: A Review},
  author={Makhataeva, Zhanat and Varol, Huseyin Atakan},
  journal={Robotics},
  volume={9},
  number={2},
  pages={21},
  year={2020},
  publisher={Multidisciplinary Digital Publishing Institute}
}
\end{document}